# A computationally and cognitively plausible model of supervised and unsupervised learning

David M. W. Powers[1,2]

[1] CSEM Centre for Knowledge & Interaction Technology, Flinders University,
Adelaide, South Australia
[2] Beijing Municipal Lab for Multimedia & Intelligent Software, BJUT
Beijing, China
powers@acm.org

**Abstract**. Both empirical and mathematical demonstrations of the importance of chance-corrected measures are discussed, and a new model of learning is proposed based on empirical psychological results on association learning. Two forms of this model are developed, the Informatron as a chance-corrected Perceptron, and AdaBook as a chance-corrected AdaBoost procedure. Computational results presented show chance correction facilitates learning.

**Keywords**: Chance-corrected evaluation, Kappa, Perceptron, AdaBoost

## 1    Introduction

The issue of chance correction has been discussed for many decades in the context of statistics, psychology and machine learning, with multiple measures being shown to have desirable properties, including various definitions of Kappa or Correlation, and the psychologically validated $\Delta P$ measures. In this paper, we discuss the relationships between these measures, showing that they form part of a single family of measures, and that using an appropriate measure can positively impact learning.

### 1.1    What's in a "word"?

In the Informatron model we present, we will be aiming to model results in human association and language processing. The typical task is a word association model, but other tasks may focus on syllables or rimes or orthography. The "word" is not a well-defined unit psychologically or linguistically, and is arguably now a backformed concept from modern orthology. Thus we use "word" for want of a better word, and the scare quotes should be imagined to be there at all times, although they are frequently omitted for readability! (Consider "into" vs "out of", "bring around" vs "umbringen".)

## 1.2 What's in a "measure"?

A primary focus of this paper is the inadequacy of currently used measures such as Accuracy, True Positive Rate, Precision, F-score, etc. Alternate chance-corrected measures have been advocated in multiple areas of cognitive, computational and physical science, and in particular in Psychology in the specific context of (unsupervised) association learning [1-3], where $\Delta P$ is considered "the normative measure of contingency".

In parallel, discontent with misleading measures of accuracy was building in Statistics [4,5], Computational Linguistics [6] and Machine Learning [7] and extended to the broader Cognitive Science community [8]. Reversions to older methods such as Kappa and Correlation (and ROC AUC, AUK, etc.) were proposed and in this paper we explore learning models that directly optimize such measures.

## 2 Informedness, Correlation & DeltaP

The concept of chance-corrected accuracy measures has been reinvented several times in several contexts, with some of the most important being Kappa variants [4,5]. This is an *ad hoc* approach that subtracts from accuracy (Ac) an estimate of the chance-level accuracy (EAc) and renormalizes to the form of a probability $K=(Ac–EAc)/(1–EAc)$. But different forms of chance estimate, different forms of normalization, and different generalizations to multiple classes or raters/predictors, lead to a whole family of Kappa measures of which $\Delta P$ turns out to be one, and $\Delta P'$ another [9]. The geometric mean of these two unidirectional measures is correlation, which is thus a measure of mean association over both directions of an A↔B relation between events. Perruchet and Pereman [3] focus on an A, B word sequence and define $\Delta P$ as a chance-corrected version of TP = P(B|A), corresponding to Precision (proportion of events A that predict B correctly), whilst $\Delta P'$ corrects TP' = P(A|B) which is better known as TPR, Sensitivity or Recall, meaning the proportion of events B that are predicted by A – on the assumption that forward prediction A→B is normative. They argue for comparing TP with a baseline of how often event B occurs when not preceded by A so that $\Delta P$ = P(B|A) – P(B|¬A) and $\Delta P'$ = P(A|B) – P(A|¬B).

Empirically $\Delta P'$ is stronger than $\Delta P$ in these experiments, and TP and TP' are much weaker, with TP failing to achieve a significant result for either Children or Adults in their experiments. Why should the reverse direction be stronger? One reason may be that an occurrence in the past is more definite for the speaker and has been more deeply processed for the hearer. Furthermore, often a following segment may help disambiguate a preceding one. Thus in computational work at both word level and phoneme/grapheme level, the preceding two units and the succeeding three units, seem to be optimal in association-based syntax and morphology learning models [10,11], and two-side context has also proven important in semantic models [12]. However, Flach [7] and Powers [8] independently derived $\Delta P'$-equivalent measures, not $\Delta P$, as a skew/chance independent measure for A→B predictions as the information value relates to (and should be conditioned on) the prevalence of B not A.

In view of these Machine Learning proofs we turn there to introduce and motivate definitions in a statistical notation that conflicts with that quoted above from the Psychology literature. We use systematic acronyms [7,8] in upper case for counts, lower case for rates or probabilities. In dichotomous Machine Learning [7] we assume that we have for each instance a Real class label which is either Positive or Negative (counts, RP or RN, rates rp=RP/N and rn=RN/N where we have N instances labelled). We assume that our classifier, or in Association Learning the predictor, specifies one Predicted class label as being the most likely for each instance (counts, PP or PN, probs pp and pn). We further define True and False Positives and Negatives based on whether the prediction P or N was accurate or not (counts, TP, TN, FP, FN; probs tp, tn, fp, fn; rates tpr=tp/rp, tnr=tn/rn, fpr=fp/rn).

Table 1: Prob notation for dichotomous contingency matrix.

|      | +R | −R |    |
|------|----|----|----|
| +P   | tp | fp | pp |
| −P   | fn | tn | pn |
|      | rp | rn | 1  |

Whilst the above systematic notation is convenient for derivations and proofs, these probabilities (probs) are known by a number of different names and we will use some of these terms (and shortened forms) for clarity of equations and discussions.

The probs rp and rn are also known as Prevalence (Prev) and Inverse Prevalence (IPrev), whilst pp and bn are Bias and Inverse Bias (IBias) resp. Also Recall and Sensitivity are synonyms for true positive rate (tpr), whilst Inverse Recall and Specificity correspond to true negative rate (tnr). The term rate is used when we are talking about the rate of finding or recalling the real item or label, that is the proportion of the real items with the label that are recalled. When we are talking about the accuracy of a prediction in the sense of how many of our predictions are accurate we use the term accuracy, with Precision (Prec) or true positive accuracy being tpa=tp/pp, and Inverse Precision or true negative accuracy being tna=tn/pn, and our (perverse) prediction accuracy for false positives being fpa=fp/pp. We also use fpa and fna correspondingly for the perverse accuracies predicting the wrong (false) class. Names for other probs [13] won't be needed.

The chance-corrected measure $\Delta P'$ turns out to be the dichotomous case of Informedness, the probability that a prediction is informed with respect to the real variable (rather than chance). This was proven based on considerations of odds-setting in horse-racing, and is well known as a mechanism for debiasing multiple choice exams [8,13]. It has also been derived as skew-insensitive Weighted Relative Accuracy (siWRAcc) based on consideration of ROC curves [7]. As previously shown in another notation, it is given by:

$$\Delta P' = tpr - fpr = tpr + tnr - 1 = \text{Sensitivity} + \text{Specificity} - 1 \qquad (1)$$

The inverse concept is Markedness, the probability that the predicting variable is actually marked by the real variable (rather than occuring independently or randomly). This reduces to $\Delta P$ in the dichotomous case:

$$\Delta P = tpa - fpa = tpa + tna - 1 = Prec + IPrec - 1 \qquad (2)$$

As noted earlier, the geometric mean of $\Delta P$ and $\Delta P'$ is Matthews Correlation (Perruchet & Pereman, 2004), and kappas and correlations all correspond to different normalizations of the determinant of the contingency matrix [13]. It is noted that $\Delta P'$ is recall-like, based on the rate we recall or predict each class, whilst $\Delta P$ is precision-like, based on the accuracy of our predictions of each label.

The Kappa interpretation of $\Delta P$ and $\Delta P'$ in terms of correction for Prevalence and Bias [9,13] is apparent from the following equations (noting that Prev<1 is assumed, and Bias<1 is thus a requirement of informed prediction, and E(Acc)<1 for any standard Kappa model):

$$\text{Kappa} = (\text{Accuracy} - E(\text{Acc})) / (1 - E(\text{Acc}))$$
$$\Delta P' = (\text{Recall} - \text{Bias}) / (1 - \text{Prevalence}) \qquad (3)$$
$$\Delta P = (\text{Precision} - \text{Prevalence}) / (1 - \text{Bias}) \qquad (4)$$

If we think only in terms of the positive class, and have an example with high natural prevalence, such as water being a noun say 90% of the time, then it is possible to do better by guessing noun all the time than by using a part of speech determining algorithm that is only say 75% accurate [6]. Then if we are guessing our Precision will follow Prevalence (90% of our noun predictions will be nouns) and Recall will follow Bias (100% of our noun occurences will be recalled correctly, 0% of the others).

We can see that these chance levels are subtracted off in (3) and (4), but unlike the usual kappas, a *different* chance level estimate is used in the denominator for normalization to a probability – and unlike the other kappas, we actually have a well defined probability as the probability of an informed prediction or of a marked predictor resp. The insight into the alternate denominator comes from consideration of the amount of room for improvement. The gain due to Bias in (3) is relative to the chance level set by Prevalence, as $\Delta P'$ can increase only so much by dealing with only one class – how much is missed by this blind 'positive' focus of tpr or Recall on the positive class is captured by the Inverse Prevalence, (1 – Prevalence).

Informedness and Markedness in the general multiclass case, with K classes and the corresponding one-vs-rest dichotomus statistics indexed by k, are simply

$$\text{Informedness} = \Sigma_k \text{Bias}_k \, \Delta P'_k \qquad (5)$$
$$\text{Markedness} = \Sigma_k \text{Prev}_k \, \Delta P_k \qquad (6)$$

Informedness can also be characterized as an average cost over the contingency table cells $c_{pr}$ where the cost of a particular prediction p versus the real class r is given by the Bookmaker odds: what you win or lose is inversely determined by the prevalence of the horse you predict (bet on) winning (p=r) or losing (p≠r) – using a programming convention for Boolean expressions here, (true,false)=(1,0), define Gain $G_{pr}$ to have

$$G_{pr} = 1/(\text{Prev}_p - D_{pr}) \qquad \text{where } D_{pr} = (p \neq r) \qquad (7)$$
$$\text{Informedness} = \Sigma_p \text{Bias}_p \, [\Sigma_r c_{pr} G_{pr}] \qquad (8)$$

Here the nested sum is equivalent to $\Delta P_p^?$ and represents how well you do on a particular horse p (a probability or pay off factor between 0 and 1). The outer sum is (5) and shows what proportion of the time you are betting on each horse.

The formulae can also be recognized in the equiprevalence case as the method of scoring multiple choice questions. With 4-horse races or 4-choice questions, all equally likely, and us just guessing, Bias = Prev = ¼, and we have three chances of losing ¼ and one of gaining ¾. We likely select the correct answer one time in four, and our expected gain is 0: ¼ / ¼ – ¾ / ¾. The odds are ¾ : ¼ but we normally multiply that out to integers so we have 3 : 1.

If we were four poker players all putting in a quarter before looking at our cards, we would have a dollar in the pool and whatever I gain someone else has lost, but my expected loss or gain is 0: 3 * ¼ + 1 * ¾. There is $1 or an Informedness of 1, at stake for every bet we make here.

## 3  Association Learning & Neural Networks

We have seen that chance-corrected ΔP measures are better models both from a statistical point of view (giving rise to probabilities of an informed prediction or marked predictor) and also from an empirical psychology perspective (reflecting human association strength more accurately). They also have the advantage over correlation of being usable separately to provide directionality or together to provide the same information as correlation. This raises the question of whether our statistical and neural learning models reflect appropriate statistics. The statistical models traditionally directly maximize accuracy or minimize error, without chance correction, and many neural network and convext boosting models can shown to be equivalent to such statistical models, as we show in this section and the next. Our question is whether these can be generalized with a bioplausble chance-correcting model.

### 3.1 The generalized Perceptron

Perceptrons (or Φ-machines) as the heart of the leading supervised neural networks, and (Linear or Kernel) Support Vector Machines as the common classifier of choice in Machine Learning, are actually equivalent models, seeking a (linear) separating boundary (hyperplane) between the positive and negative examples. If the examples are indeed linearly separable (or we can find an appropriate non-linear kernel to separate them), then SVM focuses on just one more example than there are dimensions in the separating hyperplane (the support vectors) in order to maximize the no-man's land between. In this case, both Perceptron and SVM will be perfect on the training data, and the early stopping margin Perceptron [14] or the SVM will actually do better on unseen data for not having tried to minimize the sum of squared error (SSE) as is effectively done when non-separable.

Multilayer Perceptrons or MLP (usually trained with some form of backpropagation) and Simple Recurrent Networks or SRN [15] are both networks of Perceptrons and inherit the SSE statistics as well as the backpropagation training method, which is acknowledged not to be particularly bioplausible [16] although attempts have been made to bridge the gap [17]. Other ways of training supervised and unsupervised networks are possible, and have been used in language learning experiments, including more complex recurrent networks [16,18]. But all these networks tend to use some variant of the Hebbian learning rule (10) – the main difference being whether update always takes place (unsupervised or association models) or takes place only under specific conditions (supervised models based on updates as correction only).

We now consider how these neural network and learning models fail to match the desired chance-corrected probability estimates and empirical association experiments, and develop an alternate model that does. We follow the same conventions that Boolean yes/no events are represented by (1,0) for (true,false), but note that many neural models use (1,-1) including MLP/BP with the tanh transfer function as f() which is argued to better balance the effort expended on positive and negative examples. However, biologically plausible networks conventionally separate out excitatory (+ve) and inhibitory (-ve) roles. On the other hand, there are issues modeling inhibition with subtraction given we assume neural activity (unlike kappas) can't go negative. We will discuss a multiplicative variant of the Perceptron shortly (Winnow), and we propose a model of synapse that is not strictly excitatory or inhibitory, but rather divisive (or facilitative) – noting that, due to the possibilities of scaling activity on both sides, the +ve/-ve distinction is moot.

### 3.2 A family of neural update rules

The Hebb [19] update rule can be characterized as "the neurons that fire together wire together" [17], with the basic neuron accumulation and update equations being shown in (9) & (10), where X is a collection of instances represented as a sequence of attribute vectors (and corresponds to a set of input neurons per attribute), and Y is a corresponding sequence of real class labels (desired output for each output neuron), while Z is the sequence of predicted class labels (actual output for each output neuron), which we show in summation form as well as in matrix form (with its omitted subscripts and implied sum over the inner subscripts):

$$Z = \theta(XW) ; \quad Z_{ik} = f(\Sigma_j \, g(X_{ij}) \, W_{jk}) \qquad (9)$$
$$W = XY; \quad W_{jk} = \Sigma_{ij} \, X_{ij} \, Y_{jk}; \quad \Delta W_{jk} = \lambda X_{ij} \, Y_{jk} \qquad (10)$$

In (9) we see two alternative formulations involving a threshold function as in the original Perceptron and a transfer function as in the Multilayer Perceptron, which can be the identity function, but is usually a smoothed 'sigmoid' variant of the threshold function to allow for a finite amplification factor for backpropagation rather than an infinitely fast change as we move infinitesimally through a threshold. We also show a function g(X) which may reflect recursively deeper layers of a MLP, or a radial basis or other transformation as used by SVMs. Voting, bagging, boosting and stacking ensembles may also be construed to obey (9) for appropriate choices of f() and g().

In (10) we see the original Hebb update rule in three forms. The first two forms are the 'batch update' versions in matrix and summation notations, whilst the third is the 'incremental' version, and also includes a learning rate $\lambda \leq 1$. This is repeated for each example, often more than one each and sometimes in random order, adding $\Delta W$ to W each time. For sparse (word to word) association learning, $W_{jk}$ simply corresponds to unnormalized $c_{jk}$ contingency table entries of (8), being normalized counts $c_{jk} = C_{jk}/N = W_{jk}/N$.

The standard Perceptron rule, by contrast, only updates if the wrong answer is given – in matrix or summation form the Boolean is again interpreted numerically and defines a matrix of binary values, whilst in incremental form either the binary or "if Boolean" interpretation can be used (no change if false):

$$W_{jk} = \Sigma_{ij} X_{ij} Y_{jk} (Y_{jk} \neq Z_{jk}); \quad \Delta W_{jk} = \lambda X_{ij} Y_{jk} (Y_{jk} \neq Z_{jk}) \quad (11)$$

The Margin Perceptron is a venerable variant of the Perceptron that has more recently been shown to have desirable margin optimization properties similar to an SVM [14]. The update rule becomes

$$W_{jk} = \Sigma_{ij} X_{ij} Y_{jk} (\gamma > |Y_{jk}Z_{jk}|); \Delta W_{jk} = \lambda X_{ij} Y_{jk} (\gamma > |Y_{jk}Z_{jk}|) \quad (12)$$

Here the parameter $\gamma$ represents the margin width, but can be set to 1 [14] if X and W are not explicitly normalized (as here). A soft modification of this variant, that takes less account of possibly noisy margin violations is

$$W_{jk} = \Sigma_{ij} X_{ij} Y_{jk} (1-|Y_{jk}Z_{jk}|); \Delta W_{jk} = \lambda X_{ij} Y_{jk} (1-|Y_{jk}Z_{jk}|) \quad (13)$$

Winnow [20] is a variant on the Perceptron that uses multiplication rather than addition to update the weights, in order to eliminate the contribution of irrelevant attributes, characterized by quotient rather than difference:

$$W_{jk} = \Pi_j (Y_{jk} \geq Z_{jk}) * \alpha; \quad QW_{jk} = (Y_{jk} \geq Z_{jk}) * \alpha \quad (14)$$

Note that where an error occurs for negative (Y=0) class member the corresponding weight is zeroed. Winnow2 is less severe and uses the reversible

$$QW_{jk} = (Y_{jk} \geq Z_{jk}) * \alpha + (Y_{jk} < Z_{jk}) / \alpha = (Y_{jk} \geq Z_{jk}) ? \alpha : \alpha^{-1} \quad (15)$$

Note too that Winnow's weight is exponential in the number of up corrected examples (14), and Winnow2 is exponential in differential counts of up vs down corrections (15). Taking the logarithm gives us a Perceptron-like algorithm that reflects Information rather than Prevalence, but Information is inverse to log(Prob) giving weight to surprise value or novelty rather than weight of numbers or ubiquity.

Often authors of neuroplausible models have the rider that cells may correspond to a cluster of neurons rather than one. We actually show cells that are explicitly clusters of neurons in Fig. 1(a), revealing exemplar shadow and mirror cells in inset (b).

### 3.3 The Informatron

To model chance-correction, we require a matrix that reflects Informedness gains (in "dollars") rather than counts (10) or errors (11-13). Considering each predictor separately, this profit matrix corresponds to the inner sum of (8) and thus

$$W_{jk} = \Sigma_j X_{ij} Y_{jk} \quad G_{jk} \quad (16)$$

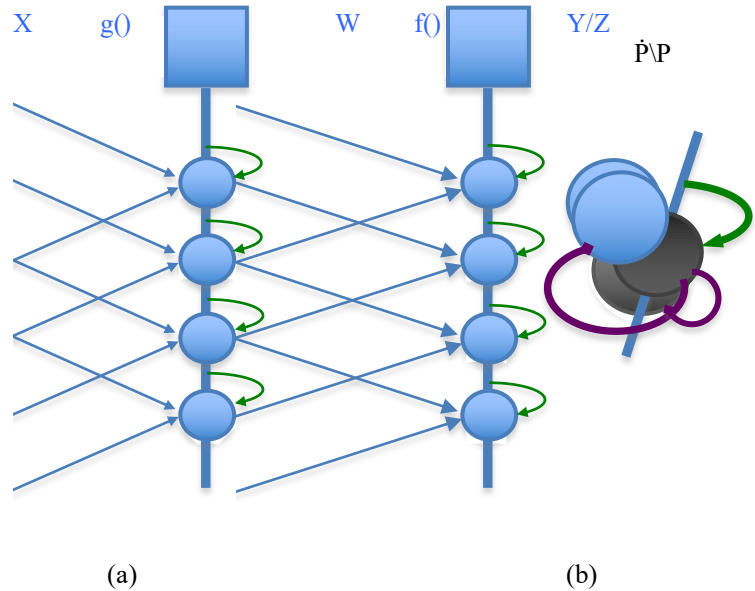

(a)                  (b)

Figure 1. The Informatron

It will be noted that no update (delta) rule is shown, although one could be if the prevalences, and hence the Gain matrix, were assumed known. We however assume that $Prev_r$ is *not* known but accumulated within the model, as shown in Fig. 1 as P.

Figure 1 shows representative synapses of a feature association or learning network to the left, corresponding to g() in (9). This is assumed to be recursively definable using the same model, which is also able to be self-organizing since it models associations between natural inputs or features, and corresponds to the perceptual and linguistic processing necessary to recognize a word from its phonological or orthological input representation. The model is thus agnostic as to whether it is unsupervised, or implicitly or explicitly supervised by feedback with or without recurrence [10,15], or may follow a similar model to the ones presented here, which is a single association stage. We make a connection to boosting and ensemble techniques here, and thus can also call it a weak learner or an individual classifier. These concepts will be picked up in the next section.

In Fig. 1(a) columns of round cells represent the before and after terms in a temporal association sequence [3]. We see here excitatory neurons obeying the standard Hebbian learning rule (10), the synapses between the columns reflecting the joint probability of the events (independent of time sequence or causality), but the simplified graphic should not be taken as precluding connections within a column – indeed the columns are shown separate only for didactic purposes and all units are activated by "words".

The square event sync cells ($X_{ek}$) synapse on all the shadow cells ($Y_{sk}$) below them with the same Hebbian learning (10), the vertical axon with curved dendritic synapses reflecting the simple (marginal) probability of the "word" events. Because they are always 1, the marginal probabilities are learned, rather than the contingencies between two concept neurons. (The square cells may be regarded in electronic engineering terms as underlying system clocks enabling a bank of cells; in cognitive neuroscience terms they may be reflected in Event Related Potentials such as the P300 and BP.)

Arrow heads represent excitatory synapses with Hebbian learning. Diamond heads represent facilitatory synapses with divisive rather than subtractive or thresholded effect, and so facilitation of the foreground neuron accumulating joint probability by the background shadow neuron accumulating marginal probability, as shown in the glide out detail of Fib. 1(b). We now show an equation corresponding to (7-8) clarifying the role of the shadow neurons:

$$Z_{ik} = f(\Sigma_j\ g(X_{ij})\ W_{jk}\ /\ S_{ik})\ \text{with}\ S_{ik} = Y_{ek} - D_{ik} \qquad (17)$$

Note that the inhibitory effect of the shadow neuron represents the normalization by prevalence of (7) & (8), but the Hebbian synaptic modification of associating foreground is independent of this gain factor. The $D_{ik}$ (which might correspond to a mismatch negativity effect and might be involved in disabling Hebbian learning and achieving Perceptron-like learning) is not illustrated for space reasons (but is a standard neural circuit involving a comparator neuron and the illustrated memory or mirror neuron, with information assumed to shift through layers at the data rate, which may also be clocked by "P300" event synchronization).

Whilst (17) is simple and reflects (8), the neural model is thus far very speculative and challenges biological plausibility with some new proposals and assumptions. Furthermore it doesn't explicitly give multiclass Informedness but that is a straightforward higher level embedding, and it doesn't model features or kernels, which is an obvious lower level recursion. We now clarify how we see the shadow and mirror neurons implementing $S_{ik}$ and give an idea of the complexity of the model suggested in Fig. 1(b).

We assume that signals shift through of the order of four layers of memory neurons, as suggested by Cohen's Magical Number Four, providing short term memory essential for associations to form and comparisons to be made, although we show only one such neuron in Fig. 1(b) as that is all that is needed for our purposes to retain the prediction. Note that all logic functions including XOR and EQV can be achieved by two layers of Perceptron-like neurons acting as NAND or NOR gates [21]. These XOR and EQV circuits correspond to our (p≠r) resp. (p=r), allowing comparison of prediction and reality in our model. We have explained how Prevalence $P_k$ is directly accumulated using standard Hebbian learning conditioned by the event clock e, as $Y_{ek}$ – and the Inverse Prevalence $\dot{P}_k = 1-P_k = \Sigma_{l\neq k}\ P_l$ can be calculated from e using the divisive operator as shown in Fig. 1(b) or accumulated by lateral synapsing of all other Prevalences similar to many famous models that actually sidestepped the question of complexity of their learning unit [18].

Given the complexity is reasonable, and is indeed reduced from O(N) to O(1) by our divisive operator, the remaining question is how parsimonious the model is. The accumulation of both contingency and prevalence information is standard Hebbian, the assumption of comparison of predictor and predicted is implicit in all the Hebbian and Perceptron rules we have considered – update depends on what happens on both sides of the synapse in all the rules (10-16). The divisive alternative to subtractive inhibition is equivalent to a single transistor and a more straightforward modulation of the signal (similar to Perceptron vs Winnow).

## 4 Fusion and Boosting

We also noted earlier that both MLPs and Boosting can also be modelled by (9), and in particular AdaBoost [22] assumes a weak learner g() and uses that to learn a strong learner in a very similar way to the Perceptron algorithms we have been considering. If the first layer of AdaBoost is a Decision Stump or Perceptron or Linear SVM, then AdaBoost corresponds to a two stage training mechanism for a two layer Perceptron. The first layer, the weak learners are trained using a standard algorithm selected to be fast rather than strong, and merely has to satisfy a weak learner criterion, namely that it can be expected with high probability to learn a classifier that will do better than chance. However, the standard algorithms define that as Error <0.5, or Accuracy >0.5, where Error is the sum of fp and fn, and Accuracy is the sum of tp and tn (Table 1), and Accuracy + Error = 1, which we abbreviate as Acc = 1 – Err.

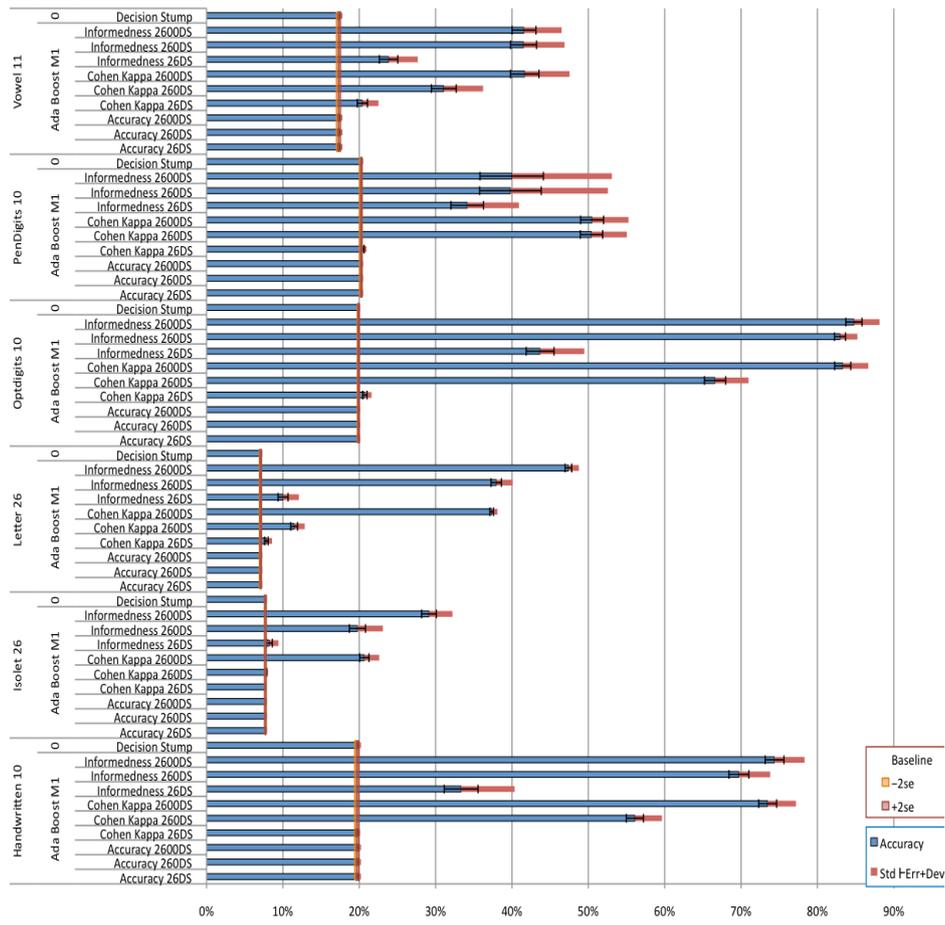

Figure 2. Accuracy of AdaBoost variants with Decision Stump weak learner.

AdaBoost sets the weight associated with each trained classifier g() to the log of the odds ratio Acc / Err, iterating while there is room for improvement (Acc < 1) and it is doing better than 'chance' (Acc > 0.5 in the standard model). Note that as Kappa = (Acc-E(Acc)) / (1-E(Acc)) goes from -1 to +1, Acc goes from 0 to 1, and the same applies for any other Kappa or Correlation measure including ΔP and ΔP' (and ROC AUC). A technique to fix this is simply to calculate $Gini_K = (K+1)/2$, where Gini (being originally designed for ROC AUC) can be applied to any chance-corrected measure K where 0 marks the chance level, mapping this chance level to ½. We can run any boosting algorithm with chance-corrected measure K by replacing Acc by $Gini_K$.

To complete the discussion of AdaBoost, it suffices to note that the different trained classifiers result from training the same weak learner on different weightings (or resamplings) of the available training set, with weights given by the odds Acc / Err.

We have now introduced a neural model that directly implements ΔP or ΔP' (which is purely a matter of direction and both directions are modelled in Fig. 1). We have also shown how a chance-corrected measure can be used for boosting, whether ΔP or ΔP' or Informedness, Markedness or Correlation, The question that follows is whether they are actually useful as learning criteria. For simplicity, we do not consider the bioplausible implementation of the neural net from this perspective, but a direct implementation of Informedness and Markedness in the context of AdaBoost.

## 5   Results & Conclusions

The most commonly used training algorithm today is SVM, closely followed by AdaBoost, which is actually usually better than SVM when SVM is boosted rather than the default Decision Stump (which is basically the best Perceptron possible based on a single input variable). To test our boosting algorithm, which we call AdaBook because of its Bookmaker corrected accuracy), we used standard UCI Machine Learning datasets relating to English letters (recognizing visually, acoustically or by pen motion). These were selected consistent with our language focus.

AdaBoost in its standard form fails to achieve any boosting on any of these datasets! AdaBook [24] with either Cohen's Kappa [4] or Powers' Informedness [8] doubles, triples or quadruples the accuracy (Fig. 2). Thus we have shown that the use of chance-corrected measures, ΔP rather than TP or TPR, etc. is not only found empirically in Psychological Association experiments, but leads to improved learning in Machine Learning experiments. This applies equally to supervised learning and unsupervised "association" learning or "clustering", and can be applied simultaneously in both directions for "coclustering" or "biclustering" [10,11,18,23].

N.B. Informedness and Information are related through Prevalence P and Euler's constant γ: $\ln P + \gamma \approx \Sigma_{p=1}^{P} 1/p$. This allows an Informatron to accumulate Information.

**Acknowledgements.** This work was supported by CNSF Grant 61070117, BNSF Grant 4122004, ARC TS0689874, DP0988686 & DP110101473, and the Importation and Development of High-Caliber Talents Project of the Beijing Municipal Institutions.